\documentclass{article}

\usepackage{arxiv}

\usepackage[utf8]{inputenc} % allow utf-8 input
\usepackage[T1]{fontenc}    % use 8-bit T1 fonts
\usepackage{hyperref}       % hyperlinks
\usepackage{url}            % simple URL typesetting
\usepackage{booktabs}       % professional-quality tables
\usepackage{nicefrac}       % compact symbols for 1/2, etc.
\usepackage{microtype}      % microtypography
\usepackage{graphicx}

\usepackage[square,numbers,sort&compress]{natbib}
\usepackage{doi}
\usepackage{multirow}

\usepackage{amsmath,amssymb,amsfonts}
\usepackage{amsthm}
\usepackage{mathrsfs}
\usepackage{bm}
\usepackage{gensymb}
\usepackage{cleveref}       % smart cross-referencing (必须放在数学包之后)

\usepackage{algorithm}
\usepackage{algorithmicx}
\usepackage{algpseudocode}
\usepackage{listings}
\usepackage{xcolor}
\usepackage{tabularx} 
\usepackage{enumerate}
\usepackage{wrapfig}
\usepackage{adjustbox}
\usepackage{bbding} 
\usepackage{array} 
\usepackage{ragged2e} 

\newcolumntype{L}[1]{>{\RaggedRight\arraybackslash}p{#1}}
\newcolumntype{C}[1]{>{\centering\arraybackslash}p{#1}}

\theoremstyle{plain}

\theoremstyle{definition}

\title{A DMD-Based Adaptive Modulation Method for High Dynamic Range Imaging in High-Glare Environments}

\date{}

\newif\ifuniqueAffiliation
\uniqueAffiliationtrue

\usepackage{authblk}

\author[1,2,*]{Banglei Guan}
\author[1,2]{Jing Tao}
\author[3]{Liang Xu}
\author[1,2]{Dongcai Tan}
\author[1,2]{Pengju Sun}
\author[1,2]{Jianbing Liu}
\author[1,2]{Yang Shang}
\author[1,2]{Qifeng Yu}

\affil[1]{College of Aerospace Science and Engineering, National University of Defense Technology, Changsha, Hunan 410073, China}
\affil[2]{Hunan Provincial Key Laboratory of Image Measurement and Vision Navigation, Changsha, Hunan 410073, China}
\affil[3]{College Of Optoelectric Science and Engineering, Zhejiang University, Hangzhou, Zhejiang 310027, China}
\affil[*]{Corresponding author: guanbanglei12@nudt.edu.cn}

\renewcommand{\shorttitle}{HDR Imaging via DMD}

\hypersetup{
	pdftitle={A DMD-Based Adaptive Modulation Method for High Dynamic Range Imaging in High-Glare Environments},
	pdfauthor={Banglei Guan et al.},
	pdfkeywords={High dynamic range imaging, Digital micromirror device (DMD), Optical modulation, Digital image correlation (DIC)},
	colorlinks=true,       % 激活颜色链接，去掉方框
	linkcolor=blue,        % 内部链接（图表、公式引用）为蓝色
	citecolor=blue,        % 参考文献引用为蓝色
	urlcolor=blue,         % 网址链接为蓝色
	anchorcolor=blue
}
\begin{document}
	\maketitle
	
	\begin{abstract}
		\textbf{Background} The accuracy of photomechanics measurements critically relies on image quality, particularly under extreme illumination conditions such as welding arc monitoring and polished metallic surface analysis. High dynamic range (HDR) imaging above 120 dB is essential in these contexts. Conventional CCD/CMOS sensors, with dynamic ranges typically below 70 dB, are highly susceptible to saturation under glare, resulting in irreversible loss of detail and significant errors in digital image correlation (DIC).
		
		\textbf{Methods} This paper presents an HDR imaging system that leverages the spatial modulation capability of a digital micromirror device (DMD). The system architecture enables autonomous regional segmentation and adaptive exposure control for high-dynamic-range scenes through an integrated framework comprising two synergistic subsystems: a DMD-based optical modulation unit and an adaptive computational imaging pipeline.
		
		\textbf{Results} The system achieves a measurable dynamic range of 127 dB, effectively eliminating saturation artifacts under high glare. Experimental results demonstrate a 78\% reduction in strain error and improved DIC positioning accuracy, confirming reliable performance across extreme intensity variations.
		
		\textbf{Conclusion} The DMD-based system provides high-fidelity adaptive HDR imaging, overcoming key limitations of conventional sensors. It exhibits strong potential for optical metrology and stress analysis in high-glare environments where traditional methods are inadequate.
	\end{abstract}
	
	
	\section{Introduction}\label{sec1}
	Optical metrology is indispensable to experimental mechanics for full-field, non-contact measurement of deformations and strains. However, its accuracy faces a fundamental limitation in high-glare environments like welding arc monitoring and polished surface analysis \cite{LI2010498,Assessment,coating,AFIFAH2025109026,RI2025108914}. Conventional CCD and CMOS sensors, with a typical dynamic range of only 60 dB, saturate under strong illumination gradients, causing irreversible information loss in highlighted regions \cite{SVE,EventHDR,High-Speed}. This severely degrades subsequent optical analyses, including digital image correlation (DIC) \cite{Digital2024,XIAO2025110999,Tensile}, grating projection profilometry \cite{LU2016103,Learning-Assisted}, and photoelastic stress measurement \cite{POGUE2018173,In-Situ}. Consequently, the limited dynamic range of imaging systems presents a major bottleneck for quantitative mechanical characterization under extreme lighting.
	
	Existing high dynamic range (HDR) methods exhibit notable limitations when applied in optical metrology. Multi-exposure fusion enhances dynamic range at the expense of temporal resolution, making it unsuitable for real-time dynamic measurements \cite{2023Multi-Exposure, ZHANG2023, LUO2025}. Global attenuation approaches, such as fixed or polarization-based neutral density filters, uniformly reduce incident light across the entire field of view. This often leads to the suppression of subtle low-intensity features and a reduction in measurement sensitivity \cite{Polarization, LI2025, Real-time}. These drawbacks emphasize the need for spatially adaptive optical modulation techniques that can preserve local image details across extreme luminance ranges without compromising speed. The proposed DMD-based system addresses this gap by achieving a measured dynamic range of 127 dB. It enables the lossless capture of extreme intensity variations and improves measurement precision in applications such as DIC.
	
\begin{figure*}[t]
		\centering
		\includegraphics[width=0.99\textwidth]{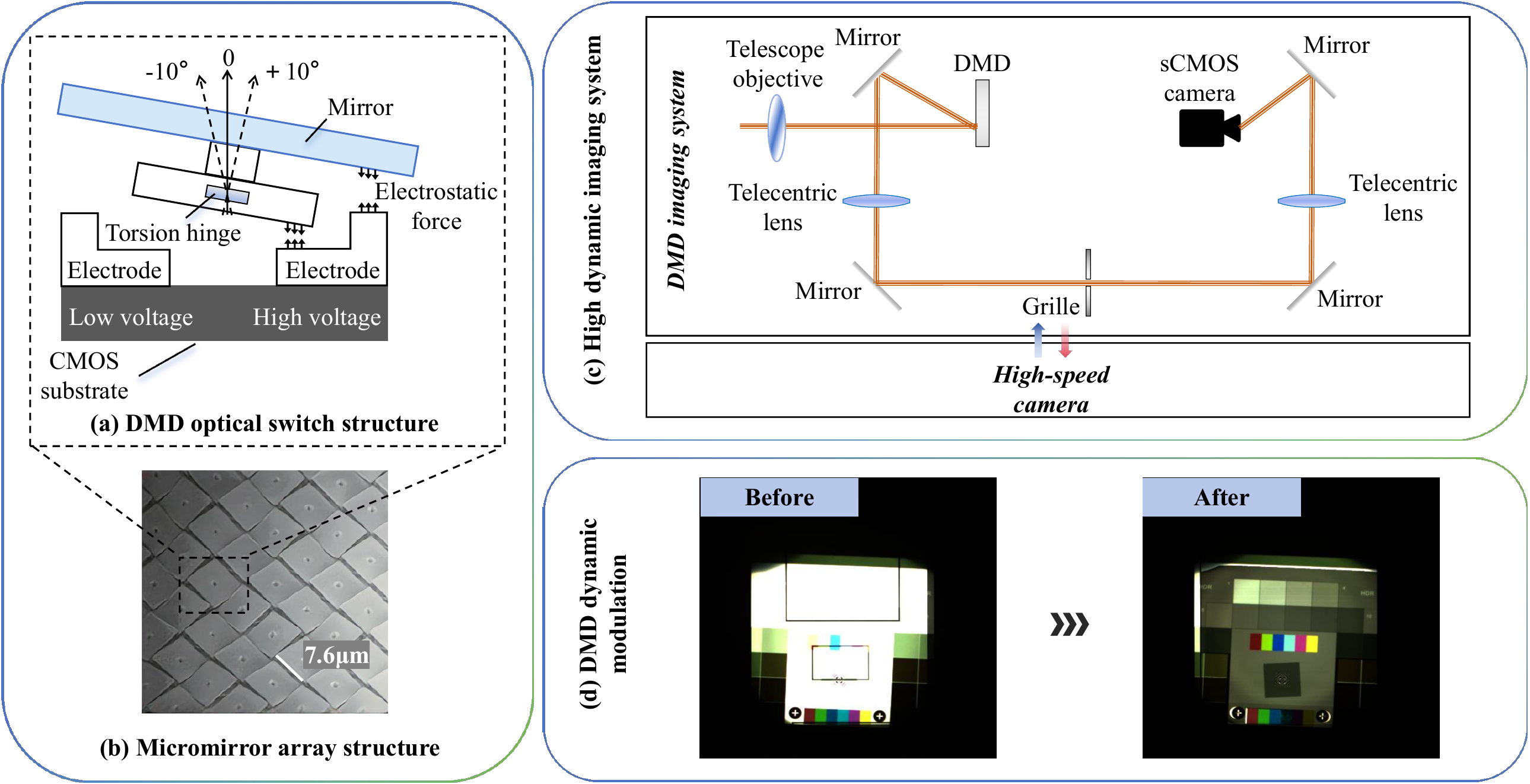}
		\caption{Schematic of the high dynamic imaging system based on DMD: (a) DMD optical switch structure; (b) Microscopic view of DMD micromirror array; (c) System setup integrating DMD modulator and high-speed camera; (d) Representative image results before and after DMD modulation.}
		\label{fig1}
	\end{figure*}

	Spatial light modulators (SLMs) have emerged as a promising solution for region-specific light control, enabling real-time, pixel-level management of incident irradiance \cite{Pixel-to-Pixel, Phase, Sun:25}. Among available SLM technologies, the digital micromirror device (DMD) is particularly suitable for optical metrology applications due to its microsecond-scale switching speed, high spatial resolution, and exceptional optomechanical stability. Originally developed by Hornbeck (1987) \cite{Hornbeck}, the DMD employs an array of individually addressable micromirrors that perform bistable tilting via binary pulse-width modulation, enabling precise grayscale light control (Fig. \ref{fig1}(a)). Nayar \emph{et al.} pioneered the use of DMDs as programmable imagers, achieving a 20 dB extension in dynamic range through spatiotemporal radiance modulation. Subsequent research by Qu \emph{et al.} adapted the DMD as a high-speed spatial shutter, facilitating high-resolution 3D shape measurement of specular surfaces \cite{Feng:17}. Recent advances in DMD technology, including 4K-resolution devices, have further enhanced their capabilities in applications ranging from projection systems \cite{Lin:14} to additive manufacturing \cite{SU2022, Progress2023} and optical aberration correction \cite{Aberration}.
	
	Despite these advancements, the potential of DMD-based adaptive modulation remains underexplored in the context of experimental mechanics, particularly for real-time HDR imaging under high-glare conditions. This paper proposes a method to enhance the detectable dynamic range of photoelectric imaging systems using a DMD. An imaging system was designed and implemented to automatically modulate light intensity from HDR targets. By integrating a DMD light intensity coding control algorithm with a target image reconstruction algorithm, high-resolution DIC measurements were achieved in extreme glare scenes. The contributions of this work are summarized as follows:
	\begin{itemize}
		\item  {Development of a novel DMD-based optical modulation system that overcomes the inherent dynamic range limitations of conventional sensors, achieving over 120 dB detectable dynamic range with spatially adaptive exposure control, unlike global attenuation filters that uniformly reduce light.}
		\item  {Introduction of an adaptive mask generation and image reconstruction algorithm that eliminates saturation artifacts through real-time localized dimming of specular reflections, addressing the temporal resolution compromises of multi-exposure fusion and the irrecoverable information loss in post-processing methods.}
		\item  {Comprehensive experimental validation demonstrating high-precision deformation tracking via DIC in high-glare environments, where conventional imaging systems typically suffer severe performance degradation and spatial contraction, thus establishing a robust solution for optical metrology under extreme illumination.}
	\end{itemize}
	
	The paper is structured as follows: Section \ref{2} delineates the system architecture, Section \ref{3} details the algorithmic framework, Section \ref{4} presents experimental validation, and Section \ref{5} concludes with implications and future research directions.
	
	\begin{figure*}[t]
		\centering
		\includegraphics[width=0.9\textwidth]{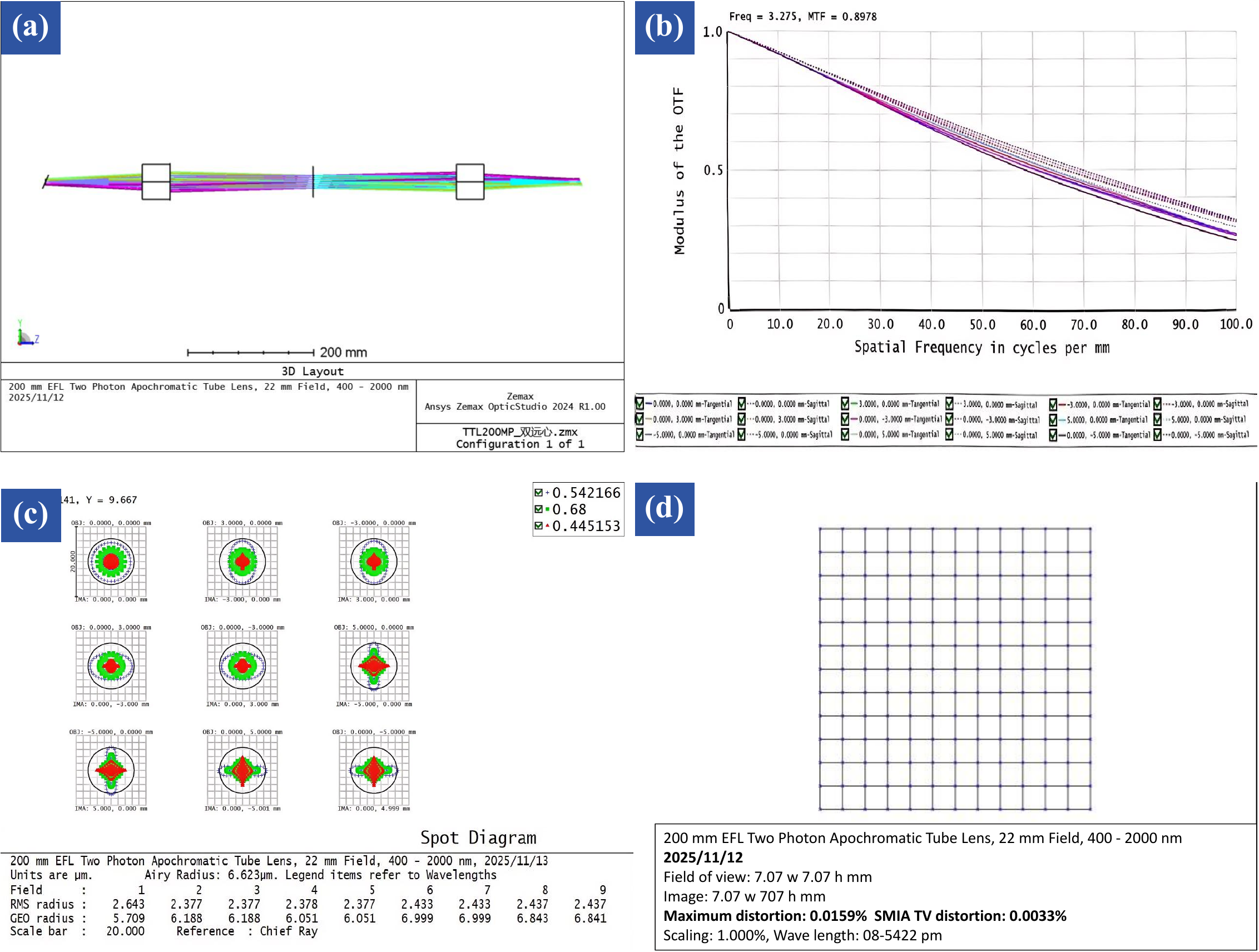}
		\caption{Zemax simulation results for the double-telecentric optical system: (a) 3D optical layout, (b) Modulation Transfer Function (MTF), (c) Spot diagram, and (d) Grid distortion ($<$ 0.1\%).}
		\label{Zemax}
	\end{figure*}
	
	\section{System Architecture}\label{2}
	To enhance the system's dynamic imaging range, this work develops a novel multi-sensor architecture (Fig. \ref{fig1}(c)). The design integrates a high-speed camera for scene analysis and grayscale segmentation, a DMD for localized intensity modulation, and a scientific CMOS (sCMOS) camera for high-dynamic-range image capture. The optical path employs a bi-telecentric lens configuration to correct aberrations and ensure measurement invariance across variable working distances \cite{2022Measurement,LI2024}.
	
	\subsection{DMD Architecture and Modulation}\label{2.1}
	The DMD, developed by Texas Instruments (TI), operates as a binary spatial light modulator based on micro-electro-mechanical principles \cite{Hornbeck}. Its core structure comprises an array of aluminum micromirrors (typically 7.56 $\mu m$ $ \times $ 7.56 $\mu m$ per mirror with 1 $\mu m$ inter-mirror gaps), each representing an independent pixel capable of $ \pm 12^\circ $ tilt around its diagonal axis at frequencies up to 9 kHz.
	The modulation mechanism relies on electrostatic actuation of individual mirrors between two states (Fig. \ref{fig1} (a)): "On" state ($ + 12^\circ $ tilt): Directs incident light toward the target optical system.
	"Off" state ($ - 12^\circ $ tilt): Diverts light away from the optical path, achieving $ > $99\% attenuation. Each mirror is addressable via an underlying CMOS memory cell, enabling binary pulse-width modulation (PWM) for precise grayscale control. This digital drive scheme eliminates phase errors and inter-pixel crosstalk, which is critical for high-fidelity spatial modulation \cite{PhysRevLett}.
	
	\begin{figure*}[t]
		\centering
		\includegraphics[width=0.90\textwidth]{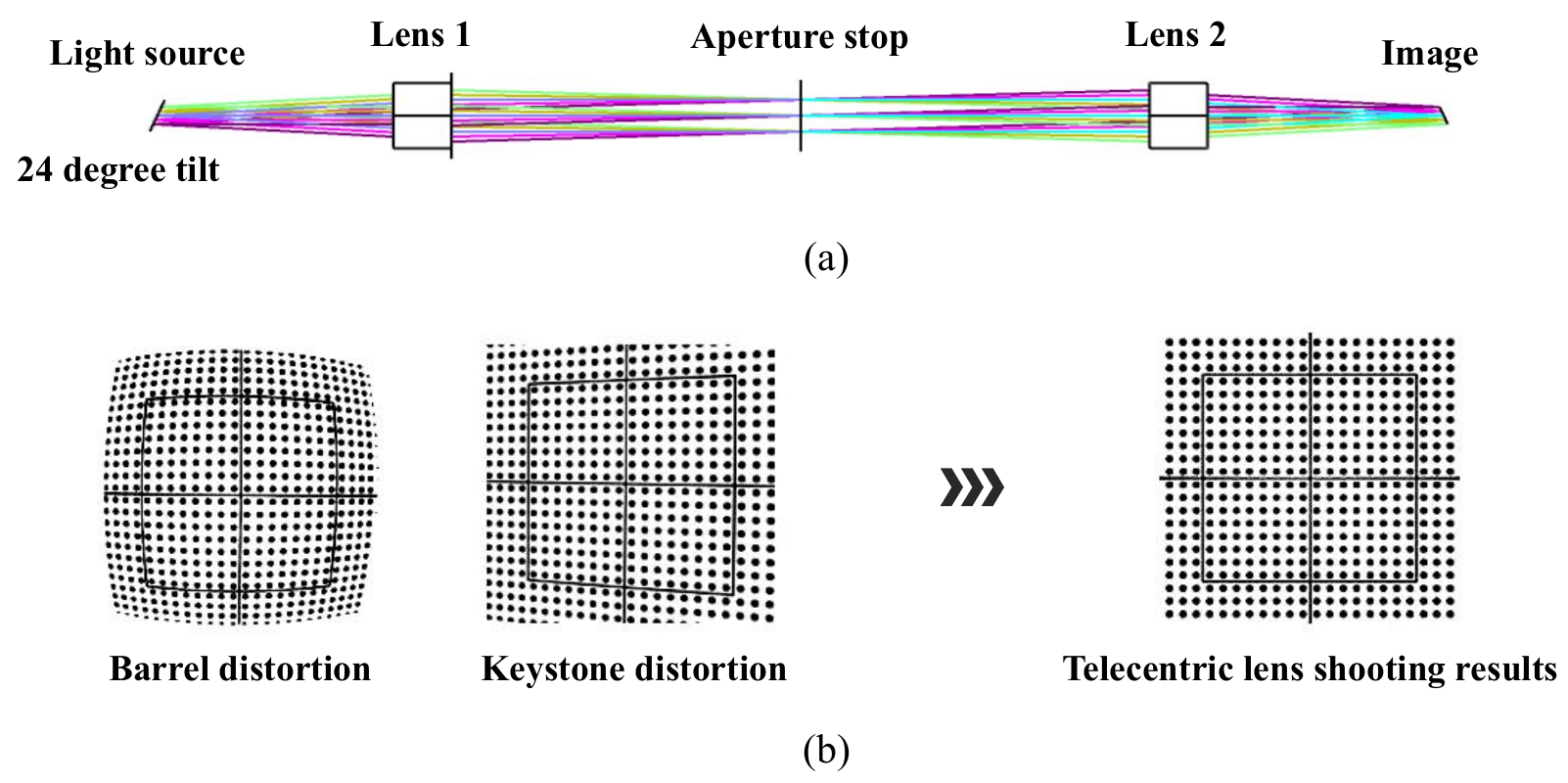}
		\caption{Double-Telecentric System Design: (a) Schematic of the double-tilted double-telecentric system on the object surface's image plane; (b) Imaging result using the double-telecentric lens.}
		\label{fig2}
	\end{figure*}
	
	\subsection{Double-Telecentric Optical Design}\label{2.2}
	Conventional non-telecentric optical systems are inadequate for modern high-precision applications due to limitations such as working-distance-dependent magnification and limited depth of field \cite{2005Handbook,HU2022106793}. These challenges are critically exacerbated in DMD-based modulation systems. Specifically, magnification variance breaks the essential pixel-to-pixel correspondence between the object, DMD, and sensor. Off-axis chief rays introduce perspective and keystone distortions---aggravated by the intrinsic $\pm 12^\circ$ mirror tilt---leading to misregistration between the computed mask and its optical realization. Furthermore, field-dependent ray angles cause non-uniform pupil fill and vignetting, which impairs the system's linear attenuation model.
	
	To overcome these multifaceted challenges, our system employs a double-telecentric design. By enforcing chief-ray parallelism on both object and image sides, it maintains constant magnification and spatial invariance across the field, preserving the pixel-level accuracy essential for high-fidelity reconstruction. This design suppresses the aforementioned geometric distortions, ensuring accurate mask registration and enhanced attenuation control. Crucially, it also enables a uniform relay of the tilted DMD image to the sCMOS sensor, stabilizing the system's optical performance despite the inherent 
	$24^\circ$ dihedral geometry. This robust optical foundation is vital for achieving the adaptive, high-fidelity HDR imaging performance our system targets.
	
	To quantitatively validate this design, we performed a Zemax simulation, with results presented in Fig. \ref{Zemax}. The simulation confirms exceptional performance: the Modulation Transfer Function (MTF) is approximately 0.55 at 60 lp/mm, indicating high resolving power for DIC speckles; the spot diagram shows well-corrected aberrations with RMS radii smaller than a pixel; and crucially, grid distortion is under 0.1\%. This low distortion directly verifies the high degree of image plane flatness essential for precision metrology. Collectively, these results provide robust evidence that the optical system delivers the high-fidelity performance required for the proposed DMD-based adaptive modulation.
	
	\subsection{DMD Optical Path Configuration}\label{2.3}
	To maximize alignment precision and optical throughput, the DMD is oriented perpendicular to the primary optical axis of the telescope objective. As illustrated in Fig. \ref{fig2}(a), incident rays parallel to this axis reflect from the DMD's micromirrors at a fixed 24\degree offset relative to the DMD's normal plane due to their intrinsic tilt geometry. Importantly, since the flip hinge of the DMD lies along its diagonal, the reflected light forms a compound angle with respect to the DMD surface. This configuration leads to two key consequences: (1) the intermediate image plane (i.e., the DMD surface) is tilted relative to the modulated optical axis, resulting in an obliquely oriented object plane at the DMD; and (2) the object plane forms a 24\degree dihedral angle with the primary optical axis.
	
	Consequently, the sCMOS detector plane must be correspondingly tilted relative to the secondary optical axis to satisfy the imaging conditions. A double-telecentric lens system (f = 200 mm) is employed to relay the tilted intermediate image to the sCMOS sensor while maintaining spatial invariance and uniform magnification across the field. This optical arrangement ensures a large depth of field and minimal distortion, both essential for high-precision imaging of tilted objects.
	
	To achieve a compact layout and accommodate the geometric constraints of the DMD, a folding mirror is strategically placed at 45\degree along the DMD diagonal. This mirror compensates for the 24\degree output angle, aligning the reflected optical path parallel to the DMD plane. The beam then propagates through two adjustable 90\degree folding mirrors before final imaging on the sCMOS. This folded design significantly reduces the system footprint compared to a straight-line configuration and incorporates multiple alignment mechanisms for enhanced flexibility. Additionally, the telescope objective is designed to be replaceable to accommodate various detection scenarios. The overall optical path configuration is depicted in Fig. \ref{fig1}(c).
	
	\section{Algorithm Design}\label{3}
\begin{figure*}[t]
		\centering
		\includegraphics[width=0.78\textwidth]{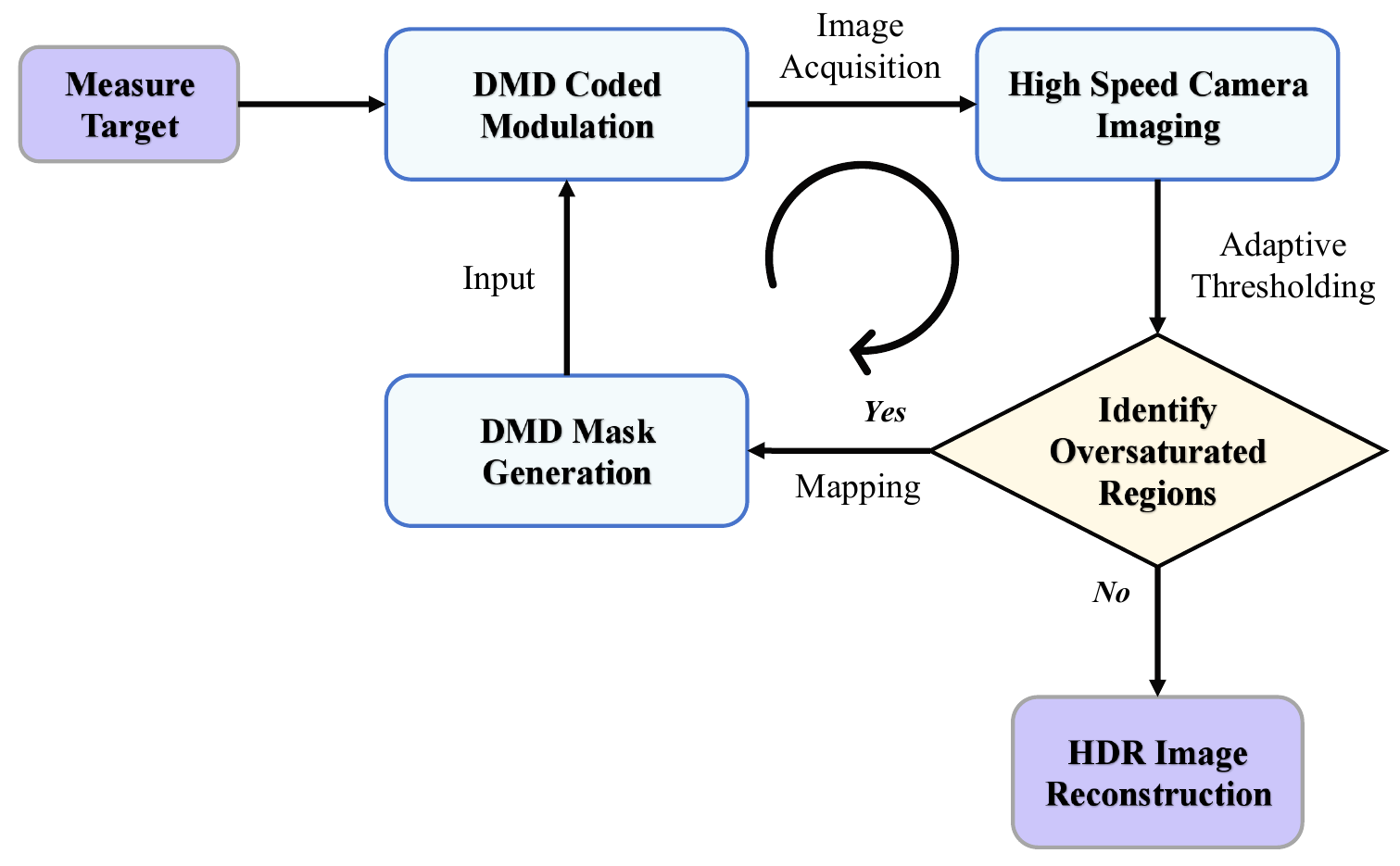}
		\caption{Flowchart of the HDR imaging process based on DMD.}
		\label{fig3}
	\end{figure*}

	This section introduces a high-speed HDR modulation algorithm for DMD-based imaging, which employs dynamic coding to achieve adaptive high dynamic range imaging. Central to this closed-loop framework is the real-time generation of modulation masks that algorithmically optimize photon distribution with micron-scale precision. The system's real-time capability, with a total latency of 15--35 milliseconds, enables an adaptive output frame rate of 28--66 Hz. This performance is achieved through a streamlined workflow: image acquisition (around 5 ms) from the sCMOS camera, mask computation (tens of ms), and near-instantaneous DMD loading via a high-bandwidth interface. Consequently, the system can dynamically adapt to varying glare conditions within a single acquisition cycle, offering a distinct advantage over conventional multi-exposure techniques. The complete workflow is illustrated in Fig. \ref{fig3}, with core components detailed in subsequent subsections.
	
	\begin{figure*}[t]
		\centering
		\includegraphics[width=0.9\textwidth]{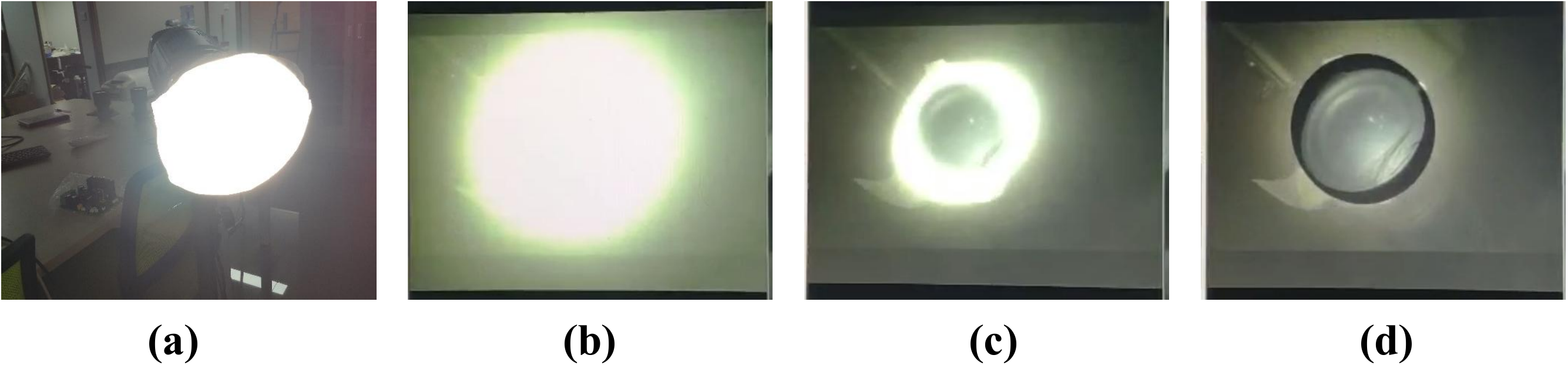}
		\caption{The light attenuation of high brightness light is realized by DMD modulation. (a) Experimental scene; (b) Image acquisition before modulation; (c) Partial suppression of images; (d) Completely suppress the image.}
		\label{fig:mask}
	\end{figure*}
	
	\subsection{Adaptive Mask Generation}\label{3.1}
	The adaptive mask generation algorithm facilitates HDR imaging through pixel-level light modulation via iterative threshold optimization. This closed-loop system dynamically attenuates saturated regions while preserving optimally exposed areas, thereby overcoming the constrained dynamic range inherent in conventional image sensors. In contrast to multi-exposure fusion techniques that are susceptible to motion artifacts in dynamic scenes, the present approach achieves artifact-free HDR imaging within a single acquisition cycle. As formalized in Algorithm \ref{alg1}, the core procedure implements:
	
	\begin{algorithm}
		\caption{Adaptive Mask Generation for DMD-based HDR Imaging}\label{alg1}
		\begin{algorithmic}[1]
			\renewcommand{\algorithmicrequire}{\textbf{Input:}}
			\renewcommand{\algorithmicensure}{\textbf{Output:}}
			
			\Require Camera saturation level $S_{\max}$ \Comment{Camera max grayscale}
			\Ensure Modulation mask $M(x,y)$ \Comment{Modulation mask}
			\State Initialize mask: $\forall$ pixels $(x,y) \gets M(x,y) = 1$ \Comment{Initialize to full-open}
			\State $S_{th} \gets 0.95 \times S_{\max}$ \Comment{Saturation threshold)}
			\State $M_{\min} \gets 2^{-11}$ \Comment{Min attenuation (1/2048)}
			\State $f \gets \text{capture\_image}(M)$ \Comment{Acquire initial frame}
			
			\While{$Z = \{ (x,y)\left| {f(x,y) > } \right.{{\rm{S}}_{th}}\} $} \Comment{Find saturated pixels}
			\If{$Z = \emptyset$ or $\min(M) = M_{\min}$} 
			\State \textbf{break} \Comment{Termination condition}
			\EndIf
			\For {$(x,y) \in Z$}
			\If{$M(x,y) > M_{\min}$}
			\State $M(x,y) \gets M(x,y) \times 0.5$  \Comment{Halve attenuation}
			\Else
			\State $M(x,y) \gets M_{\min}$  \Comment{Clamp to min}
			\EndIf
			\EndFor
			\State $f \gets \text{capture image}(M)$  \Comment{Update image}
			\EndWhile 
			\Return $M(x,y)$
		\end{algorithmic}
	\end{algorithm}
	
	This adaptive mask generation constitutes the fundamental mechanism for high-fidelity HDR imaging by dynamically optimizing photon transmission at each spatial coordinate. The resultant mask $M(x,y)$ encodes scene-specific attenuation parameters, effectively compressing extreme luminance variations into the sensor's linear response range. By selectively suppressing specular highlights while preserving normal brightness details, this approach addresses the limitation of conventional sensors in capturing both high-intensity reflections and low-illumination details.
	
	\subsection{Intensity Modulation and High Dynamic Range Synthesis}\label{3.2}
	Building upon the adaptive attenuation map $M(x,y)$ derived in Section \ref{3.1}, this module executes spatial intensity modulation and HDR synthesis. Scene radiance is projected onto the DMD plane through an objective lens, where micromirrors adopt bistable orientations: $ + 12^\circ $ (``ON'' state) directs light toward the imaging sensor, while $ - 12^\circ $ (``OFF'' state) diverts illumination to the beam dump (Fig. \ref{fig1}(a)). Through hardware-software co-design, the DMD controller dynamically configures micromirror arrays via binary bit-plane loading, while the host computational unit calculates optimal exposure sequences. Intensity modulation is achieved through temporal coded exposure, formalized as:
	\begin{equation}
		V(x,y) = \sum\limits_{i = 1}^N {\sum\limits_{t = 0}^T {{\mu _i}(x,y,t){t_i}\left[ {{{\left| {L(s,t)} \right|}^2} \otimes {{\left| {h(x,y;s,t)} \right|}^2}} \right]} } ,0 \le V \le 255
	\end{equation}
	where $\mu_i(x,y,t)$ denotes the instantaneous spatial modulation coefficient at coordinate $(x,y,t)$ (governed by DMD patterns), and $t_i$ represents the temporal weighting coefficient. The term $\left[ |L(s,t)|^2 \otimes |h(x,y;s,t)|^2 \right]$ characterizes the optical transfer convolution between the scene's intrinsic radiance distribution and the system's point spread function. Here, $N$ signifies spatial sampling points and $T$ the temporal integration limit. The HDR imaging model is mathematically formalized as:
	\begin{equation}
		{V_{{\rm{HDR}}}}(x,y) = \mu (x,y) \cdot {I_{{\rm{HDR}}}}(x,y)
	\end{equation}
	where $V_{\mathrm{HDR}}(x,y)$ corresponds to the intensity captured by the CMOS sensor, and $I_{\mathrm{HDR}}(x,y)$ denotes the true HDR radiance of the scene. Original HDR scene radiance is reconstructed through inverse modulation:
	\begin{equation}
		I_{\text{HDR}}(x,y) = \frac{V_{\text{HDR}}(x,y)}{\mu (x,y)}
	\end{equation}
	
	Conventional 8-bit sensors exhibit a fundamental dynamic range limitation:
	\begin{equation}
		R_D = 20 \cdot \lg \left( \frac{I_{\text{max}}}{I_{\text{min}}} \right) \approx 20 \cdot \lg (255) = 48.13\ \text{dB}
	\end{equation}
	
	In our system, we adopt a higher-dynamic-range 16-bit sCMOS detector, and further extend the usable range using a DMD as a programmable optical attenuator.
	Functioning as a programmable optical attenuator, the DMD primarily modulates incident light intensity. Let $T_{\max}$ and $T_{\min}$ denote the maximum and minimum micromirror ``ON'' state durations, respectively. The corresponding system-level dynamic range can be expressed by augmenting the intrinsic detector term with the DMD temporal modulation factor:
	\begin{equation}
		R_D = 20 \cdot \lg \left( \frac{I_{\text{max}}}{I_{\text{min}}} \cdot \frac{T_{\text{max}}}{T_{\text{min}}} \right)
	\end{equation}

	In our implementation, the intrinsic sCMOS dynamic range $\big(20\lg(I_{\rm max}/I_{\rm min})\big)$ is measured to be 87 dB\footnote{The sCMOS camera used in our system is a 16-bit \texttt{PCO.panda~4.2}.}.
	By configuring the temporal modulation ratio $T_{\rm max}/T_{\rm min}$ to 100, the DMD provides an additional modulation gain of 40 dB, resulting in a total dynamic range of 127 dB. Through this adaptive modulation strategy, the system selectively attenuates overexposed regions while maintaining accurate measurement in areas with normal illumination. As a result, the system delivers high-fidelity imaging under extreme luminance conditions completely free from saturation.
	
	To facilitate precise control of the modulation process, a dedicated software interface has been developed (Fig. \ref{fig:mask}). This tool allows for the dynamic adjustment of key parameters, including the attenuation level for managing dynamic range, the Gamma curve for optimizing contrast, and the gain for enhancing sensitivity. Additionally, the interface features a manual suppression mode, enabling users to selectively attenuate specific regions of interest. This flexibility enables optimized imaging performance across complex and extreme scene conditions.
	
	\begin{figure*}[t]
		\centering
		\includegraphics[width=0.8\textwidth]{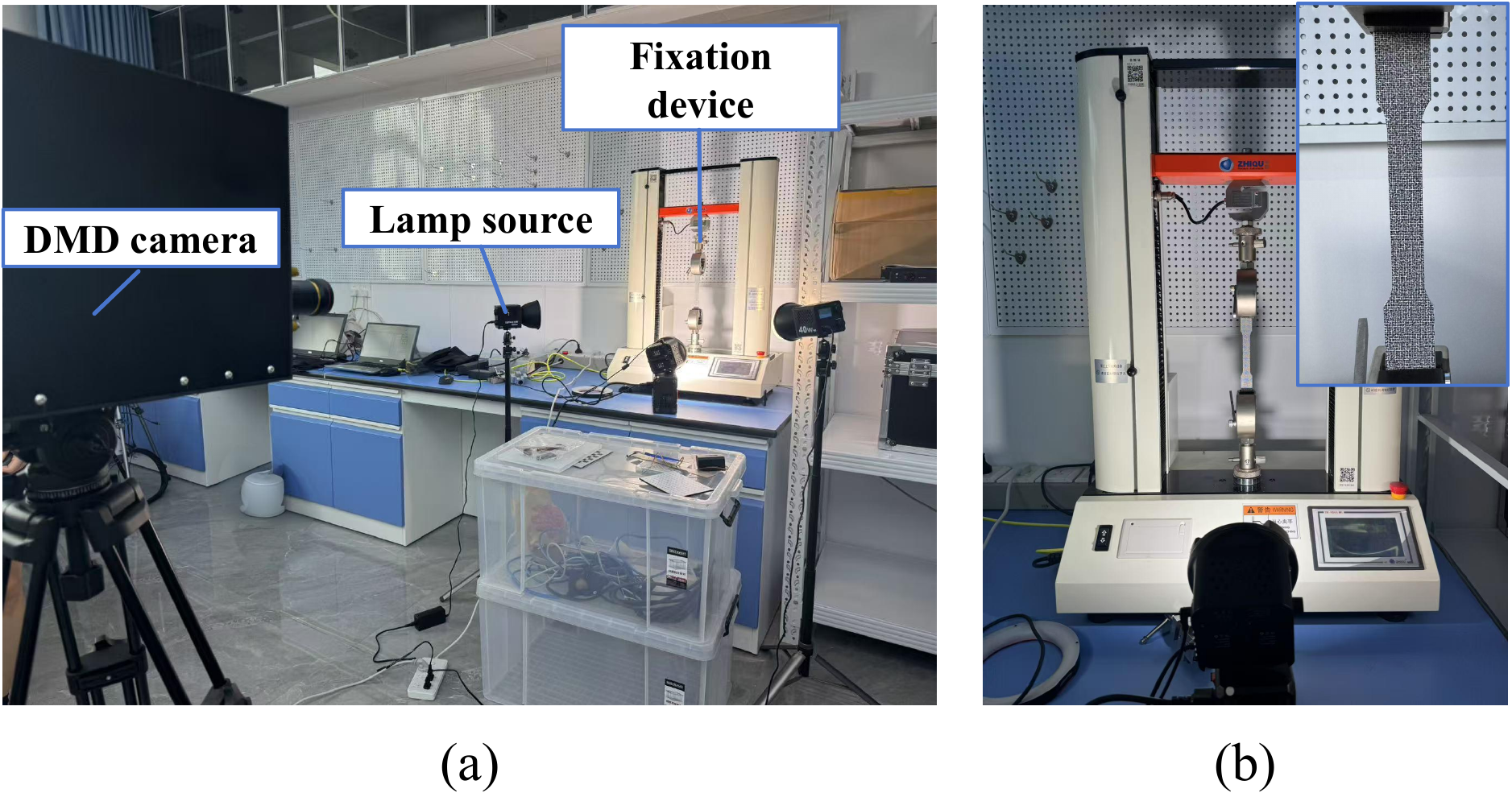}
		\caption{A schematic diagram of the experimental setup is shown. (a) The implementation diagram of the experimental scene; (b) A detailed image of the object, sprayed with a speckle pattern.}
		\label{fig4}
	\end{figure*}
	
	\renewcommand{\arraystretch}{1.3} 
	
	\begin{table*}[htbp]
		\centering
		\caption{Comparative analysis of HDR systems against critical requirements for DIC. The symbol $N/A$ indicates "Not Applicable".}
		\label{tab:HDR_device} 
		\footnotesize
		\resizebox{\textwidth}{!}{%
			\begin{tabular}{L{3.5cm} C{1.8cm} C{1.8cm} C{1.8cm} C{1.8cm} C{1.8cm}} 
				\toprule
				\textbf{} & \textbf{Computational HDR \cite{Subexposures/s}} & \textbf{Spatially Patterned Sensor \cite{Demosaicing}} & \textbf{Native Logarithmic Sensor \cite{BOUVIER2014146} } & \textbf{Event Camera \cite{event} } & \textbf{Our DMD System} \\
				\midrule
				\textbf{Full Spatial Resolution} & \Checkmark     &                &                &                & \Checkmark \\
				\textbf{Immunity to Motion Artifacts} &                & \Checkmark     & \Checkmark     & \Checkmark     & \Checkmark \\
				\textbf{Data Integrity in Glare Regions} &                &                & \Checkmark     & \Checkmark     & \Checkmark \\
				\textbf{High SNR \& Contrast} & \Checkmark     &                &                & N/A            & \Checkmark \\
				\textbf{DIC Software Compatibility} & \Checkmark     & \Checkmark     & \Checkmark     &                & \Checkmark \\
				\textbf{Core Weakness} & Motion Artifacts & Resolution Loss & Low Contrast & Data Incompatibility & None \\
				\bottomrule
			\end{tabular}
		}
	\end{table*}
	
	\section{Experiments}\label{4}
	To rigorously evaluate the efficacy of the proposed DMD system under extreme illumination conditions, the experimental configuration illustrated in Fig. \ref{fig4} is implemented. A tensile deformation scenario is designed with an auxiliary light source to simulate high-intensity glare. The evaluation comprises two principal components: (1) HDR imaging performance analysis and (2) comparative mechanical analysis. Detailed experimental procedures are delineated in subsequent subsections.
	
	\subsection{HDR Imaging Performance Evaluation}\label{4.10}
	This section evaluates the HDR imaging performance of the proposed DMD-based system. In view of the application-specific trade-offs inherent in HDR hardware design, we focus our evaluation on metrics most relevant to DIC, rather than general imaging specifications. As summarized in Table~\ref{tab:HDR_device}, conventional hardware-based HDR methods typically suffer from inherent limitations---such as motion artifacts, resolution loss, and reduced contrast---that adversely affect measurement accuracy.
	
	\begin{figure*}[t]
		\centering
		\includegraphics[width=0.93\textwidth]{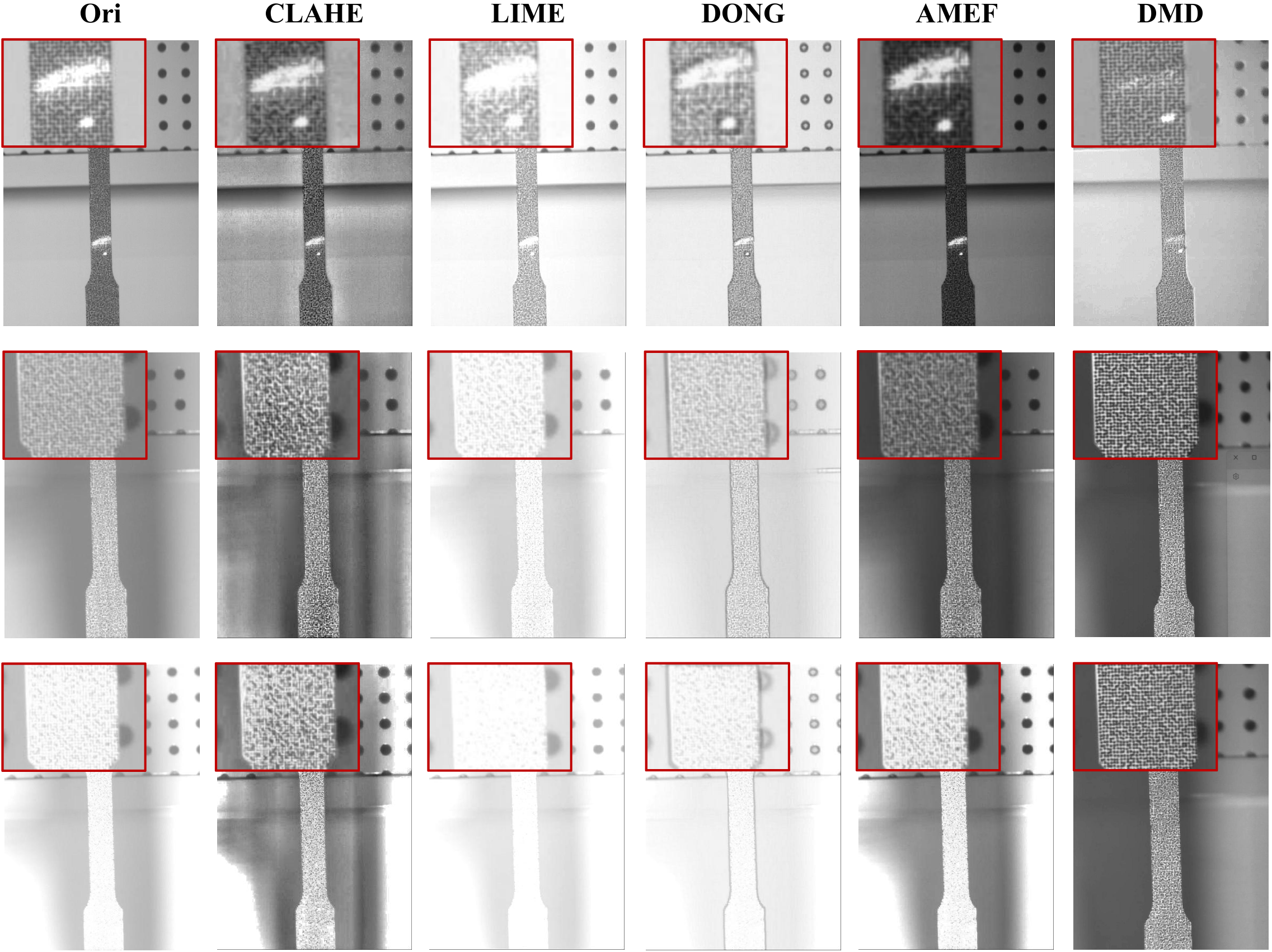}
		\caption{Visualization results comparison. Each row depicts a different light environment: first row (local overexposure, laser simulation); second row (low-level global overexposure); third row (high-level global overexposure). Each column shows a processing method: first column (original); second to fifth columns (comparison methods); sixth column (our DMD modulation result). The visual results shown are representative images from the respective experimental conditions.}
		\label{data1}
	\end{figure*}
	\begin{table}[tp]
		\centering
		\caption{Quantitative assessment using different metrics. The values presented are averaged from a sequence of multiple images for each method and scenario. Best values are in bold and second-best are underlined.}
		\label{table1}
		\begin{tabular*}{\textwidth}{@{\extracolsep\fill}ccccccc}
			\toprule
			& ORI   & CLAHE \cite{CLAHE} & LIME \cite{LIME}  & DONG \cite{dong}  & AMEF \cite{AMEF}  & DMD \\
			\midrule
			AG $\uparrow$    & 3.533 &  \textbf{7.476} & 3.799 & 4.207 & 4.801 & \underline{5.915} \\
			ENTROPY $\uparrow$ & 6.268 & \underline{6.814} & 4.998 & 5.419 & 6.552 & \textbf{6.845} \\
			CONTRAST $\uparrow$ & 0.135 & \underline{0.171} & 0.114 & 0.102 & 0.168 & \textbf{0.181} \\
			NIQE $\downarrow$  & 4.381 & 4.385 & 4.675 & 4.252 & \underline{3.788} & \textbf{3.730} \\
			BRISQUE $\downarrow$ & 41.844 &  \underline{24.785} & 33.170 & 27.232 & 25.335 & \textbf{18.972} \\
			
			\bottomrule
		\end{tabular*}
		\label{tab:table1} 
		\footnotetext{The symbol $\uparrow$ indicates that higher metric values are better, while $\downarrow$ indicates that lower values are better.}
	\end{table} 
	
	\subsubsection{Comparison with Image Enhancement Algorithms}
	We first compare our system against four state-of-the-art image enhancement algorithms: CLAHE~\cite{CLAHE}, LIME~\cite{LIME}, DONG~\cite{dong}, and AMEF~\cite{AMEF}. All comparisons use the original authors' publicly available code with default parameters. Evaluations are conducted under a range of exposure conditions, including both localized and global overexposure. This comparison underscores a key advantage of our approach: by modulating light at the optical source, it prevents irreversible sensor saturation, a limitation that post-processing methods cannot overcome.
	
	Results presented in Fig. \ref{data1} demonstrate that conventional processing techniques only partially attenuate illumination non-uniformity and show limited efficacy in reconstructing saturated regions (Row 1). This limitation stems from the inherent irrecoverability of information in overexposed areas through post-processing. Although CLAHE and AMEF achieve notable speckle restoration, they simultaneously introduce pronounced artifacts. In contrast, the proposed DMD system maintains superior visual fidelity with consistent performance across diverse luminance conditions, demonstrating robust operation in most extreme visible-light environments. It is important to note that the system remains susceptible to laser-induced glare under localized overexposure from high-energy pulse irradiation.
	
	Quantitative evaluation is performed using five objective metrics: Average Gradient (AG) \cite{CUI2015199} for sharpness assessment, Entropy for information capacity evaluation, Contrast for intensity differentiation, and two no-reference perceptual quality indices---NIQE \cite{NIQE} and BRISQUE \cite{6272356}. For each scenario, the reported metrics in Table \ref{table1} represent the average values obtained from a sequence of multiple images to ensure statistical robustness. Higher values indicate better performance for AG, Entropy, and Contrast, whereas lower values reflect superior quality for NIQE and BRISQUE. As shown in Table \ref{table1}, the proposed DMD method demonstrates statistically significant superiority, achieving the best scores in Entropy (6.845), Contrast (0.181), NIQE (3.730), and BRISQUE (18.972). These results validate its ability to effectively preserve structural details while maintaining a natural visual appearance. Although CLAHE produces the highest AG value (7.476), it does so at the cost of introducing noticeable artifacts (Fig. \ref{data1}), which accounts for its poorer perceptual scores (NIQE: 4.385; BRISQUE: 24.785). Notably, the DMD method attains a well-balanced AG value (5.915, second highest), effectively avoiding over-enhancement artifacts while retaining strong adaptability to challenging illumination conditions, thus confirming its practical utility.
	
	\begin{figure*}[t]
		\centering
		\includegraphics[width=0.92\textwidth]{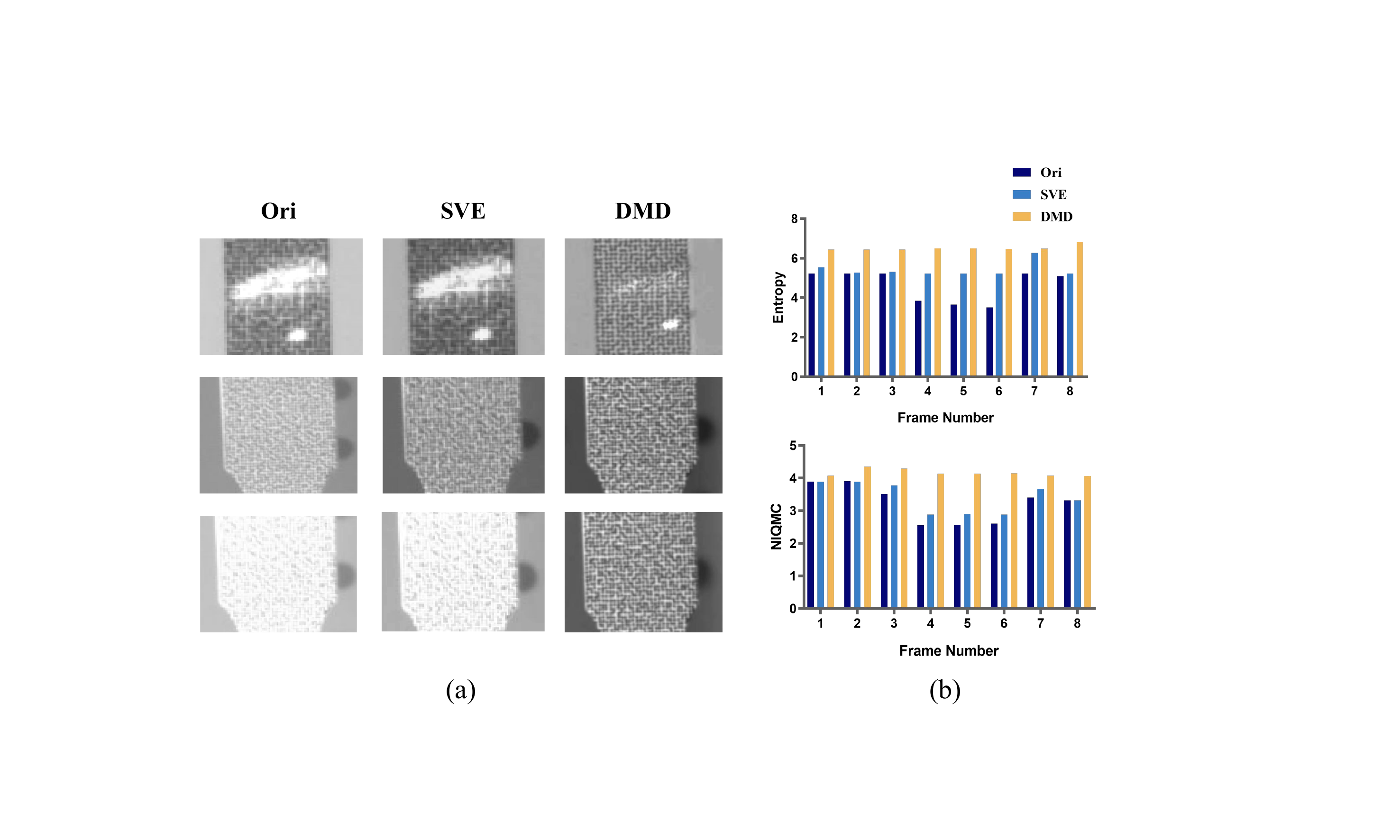}
		\caption{Comparison with a SVE camera. (a) Visual results under three glare conditions (rows): local overexposure, low- and high-level global overexposure; columns show original (Ori), SVE output, and DMD output. (b) Quantitative results: Entropy and NIQMC across eight frames (higher is better for both metrics).}
		\label{sve_dmd}
	\end{figure*}
	
	\subsubsection{Comparison with Hardware HDR Methods}
	To further validate our system's performance relative to existing hardware HDR solutions, we conducted comparative experiments with a Spatially Varying Exposure (SVE) camera~\cite{TAO2025113870,SVE}, which belongs to the "Patterned Sensor" category in Table~\ref{tab:HDR_device} and employs spatial modulation principles. Fig.~\ref{sve_dmd} presents both visual and quantitative comparisons under varying illumination conditions.
	
	As shown in Fig.~\ref{sve_dmd}(a), while the SVE camera provides some glare reduction, our DMD system achieves superior detail preservation and more effective saturation suppression across all tested scenarios. The quantitative analysis in Fig.~\ref{sve_dmd}(b) corroborates these observations: the DMD system consistently maintains higher Entropy values (averaging 6.5--7.0) compared to both original images and SVE outputs, indicating superior information preservation. Similarly, NIQMC scores demonstrate that our system maintains better perceptual quality (averaging 4.0--4.5) across all test frames.
	
	A key distinction lies in system integration. SVE cameras require subsequent fusion algorithms to combine spatially varying exposures, which can introduce artifacts and impact the effective resolution of the final HDR output. In contrast, our DMD system performs adaptive optical modulation directly at the source, delivering high-fidelity HDR images without post-acquisition fusion complexities. This integrated approach makes our DMD system particularly suitable for high-precision DIC applications where both dynamic range and image fidelity are paramount.
	
	\begin{figure*}[t]
		\centering
		\includegraphics[width=0.98\textwidth]{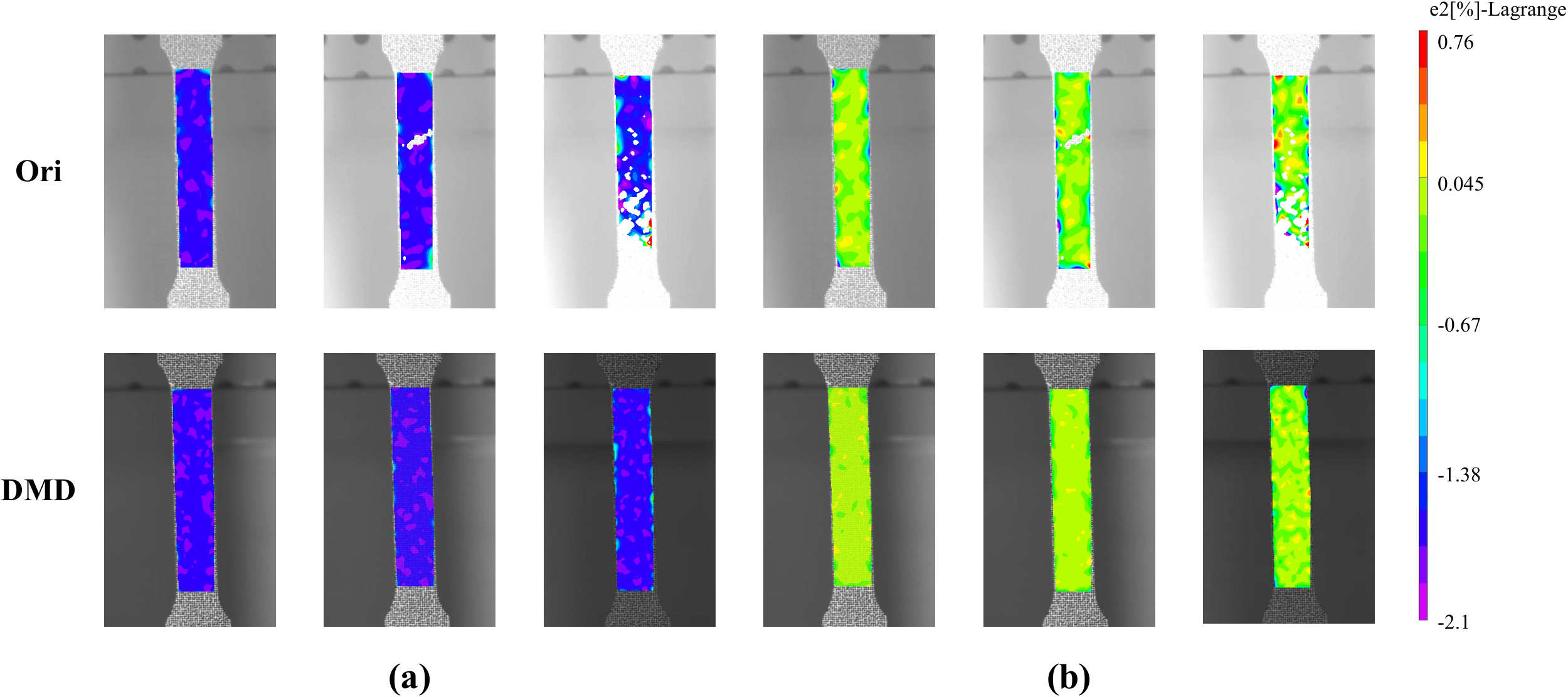}
		\caption{Visualization of static stress analysis using DIC. (a) and (b) depict computed principal stresses ${E_{\rm{1}}}$ and ${E_{\rm{2}}}$, respectively. Each panel shows increasing exposure levels from left to right. Top row: images acquired with a conventional industrial camera; bottom row: images acquired using DMD suppression. Color gradients indicate stress values (theoretical strain = 0). Cold tones represent small stress errors; warm tones represent large stress errors. The stress maps are derived from the average of a sequence of multiple images for each exposure level, ensuring robustness in the DIC analysis.}
		\label{fig5}
	\end{figure*}
	
	\subsection{Comparative Mechanical Analysis}\label{4.20}
	This section evaluates the mechanical measurement capabilities of the DMD system using two complementary static DIC experiments, where ambient light intensity serves as the controlled variable: (1) analysis of effective computational regions under extreme illumination, and (2) static strain quantification. For each exposure level, a sequence of multiple images are acquired sequentially, and the reported results in Table \ref{tab3} represent the average of these measurements to ensure robustness. The effective computational area is defined as the percentage of the successfully computed deformation field area relative to the total selected area. The main findings are summarized as follows:
	
	\begin{table}[tp]
		\centering
		\caption{Quantitative analysis of the static test results. The values presented are averaged from a sequence of multiple images for each exposure level. Best values are highlighted in bold.}
		\begin{tabular*}{\textwidth}{@{\extracolsep\fill}ccccc}
			\toprule
			Exposure level & Mode  & ${E_{\rm{1}}}$(\%) & ${E_{\rm{2}}}$(\%) & Effective calculation area(\%)\\
			\midrule
			\multirow{2}[1]{*}{1} & Ori   & 0.192 & -0.16 & 100 \\
			& DMD   & \textbf{0.073} & \textbf{-0.082} & \textbf{100} \\
			\multirow{2}[0]{*}{2} & Ori   & 0.199 & -0.218 & 97.17 \\
			& DMD   & \textbf{0.081} & \textbf{-0.085} & \textbf{100} \\
			\multirow{2}[1]{*}{3} & Ori   & 0.4   & -0.332 & 80.69 \\
			& DMD   & \textbf{0.088} & \textbf{-0.072} & \textbf{100} \\
			\bottomrule
		\end{tabular*}%
		\label{tab3}%
	\end{table}%
	
	As illustrated in Fig. \ref{fig5}, the DMD system demonstrates significantly enhanced mechanical measurement performance under extreme illumination compared to conventional imaging techniques. Progressive elevation of exposure levels induces substantial degradation in traditional methods, characterized by two correlated failure modes: spatial propagation of high-error artifacts and progressive contraction of valid measurement areas-both attributable to irreversible data loss in sensor-saturated regions. In contrast, the DMD methodology enables comprehensive boundary-to-boundary analysis across all exposure levels through dynamic optical modulation. This approach ensures consistent mapping of stress distributions while actively suppressing localized errors via real-time glare compensation. Crucially, the system maintains full-field measurement integrity even under critical light intensity overload, demonstrating exceptional robustness through stable spatial coverage and fidelity to structural boundaries. The DMD platform's capacity to sustain measurement precision and prevent performance degradation establishes it as a novel solution for reliable strain characterization in high-intensity optical environments where conventional techniques exhibit inherent limitations.
	
	Quantitative data presented in Table \ref{tab3} corroborate the DMD system's superior performance under escalating exposure. Traditional imaging exhibits progressive deterioration, manifested by: (1) expansion of strain measurement errors (${E_{\rm{1}}}$ increased by 108\% to 0.40\%; ${E_{\rm{2}}}$ deteriorated by 107\% to -0.332\% at Exposure Level 3), and (2) irreversible reduction of valid measurement areas (effective area decreased by 19.31\%). Conversely, the DMD approach maintains analytical integrity across all exposure conditions, delivering precise strain quantification (${E_{\rm{1}}}$ restricted to merely 0.088\%, ${E_{\rm{2}}}$ to -0.072\% at Level 3---representing a 78\% reduction in error compared to conventional methods) while preserving 100\% of the effective calculation area. The system thus sustains full-field measurement fidelity under extreme illumination, validating its robustness through uncompromised spatial resolution and boundary definition.
	
	\section{Conclusion}\label{5}
	This paper establishes that conventional imaging systems fundamentally fail to provide reliable mechanical measurements in high-glare environments due to irreversible sensor saturation. To overcome this critical limitation, we developed a novel DMD-based HDR imaging system featuring autonomous regional segmentation and adaptive exposure control. Our integrated framework, synergistically combining dynamic optical modulation with computational imaging, achieves three transformative advancements: (1) extending the detectable dynamic range beyond 120 dB by eliminating highlight saturation while preserving critical shadow details, enabling artifact-free imaging under extreme luminance gradients; (2) significantly enhancing measurement robustness, evidenced by quantitative DIC experiments confirming a 78\% reduction in strain errors and the crucial complete retention of valid measurement areas under critical illumination, thereby overcoming the irreversible performance degradation of conventional CCD/CMOS systems; and (3) achieving spatially adaptive precision, validated through boundary-to-boundary analysis at micron-scale resolution that maintains structural fidelity even at the edges of high-contrast regions and eliminates the spatial contraction and error propagation endemic to traditional techniques. Collectively, these capabilities establish the DMD platform as a transformative solution for mechanical quantification in high-intensity optical environments such as welding monitoring and polished surface analysis. Future work will focus on computational acceleration for real-time applications and system miniaturization for portable field deployment.
	
	\section*{Acknowledgements}
	This work was supported by the National Natural Science Foundation of China (Grant No. 12372189).
	
	\section*{Data availability}
	Data underlying the results presented in this article are not publicly available at this time, but may be obtained from the authors upon reasonable request.
	
	\section*{Declarations}
	\textbf{Conflict of Interest} The authors declare that they have no conflicts of interest.
	
	\bigskip
	
	\bibliographystyle{unsrtnat}
	\bibliography{ref.bib}
	
\end{document}